%% file: bio-inspired.tex
\documentclass[conference]{IEEEtran}
\IEEEoverridecommandlockouts
\usepackage{cite}
\usepackage{amsmath,amssymb,amsfonts}
\usepackage{algorithmic}
\usepackage{graphicx}
\usepackage{textcomp}
\usepackage{xcolor}
\usepackage{booktabs}
\usepackage[hidelinks]{hyperref}
\usepackage{multirow}
\usepackage{orcidlink}

\def\BibTeX{{\rm B\kern-.05em{\sc i\kern-.025em b}\kern-.08em
    T\kern-.1667em\lower.7ex\hbox{E}\kern-.125emX}}

\begin{document}

\title{The Whale That Outswam Evolution: Swarm Intelligence Maximises
Memory in Connectome Reservoirs%
\thanks{Corresponding author: A.~Guragain (anmol.g@upm.es).
Code and data: \url{https://github.com/Anmol2059/connectome-reservoir-optimisation}}%
}

\author{%
\IEEEauthorblockN{Anmol Guragain\,\orcidlink{0000-0000-0000-0000},
                  Savvas Kakalis, and
                  Juan Ignacio Godino-Llorente}
\IEEEauthorblockA{%
  \textit{ETSI de Telecomunicaci\'{o}n},
  Universidad Polit\'{e}cnica de Madrid,
  Madrid, Spain}
}

\maketitle

\begin{abstract}
Reservoir computing exploits the fixed dynamics of a recurrent network
for temporal processing, requiring only a trained linear readout.
Biological neural connectomes, shaped by millions of years of evolution,
may encode computational structure beyond what random reservoirs provide,
yet whether that structure can be further enhanced by principled optimisation
remains an open question.
We address it by applying four gradient-free, bio-inspired optimisers
(Particle Swarm Optimisation, Differential Evolution, Grey Wolf Optimiser,
and Whale Optimisation Algorithm) to the edge weights of
connectome-based echo-state networks across six species spanning six orders
of magnitude in neural complexity: \textit{C.~elegans} (279 neurons),
\textit{Drosophila} (49 nodes), mouse (112), rat (73), macaque (29 regions,
continuous FLNe synaptic strengths), and human structural MRI connectivity
(83 parcels).
Each connectome is evaluated on four canonical reservoir computing benchmarks:
Memory Capacity (MC), Lorenz attractor prediction, NARMA-10 system
identification, and Mackey--Glass chaotic time-series prediction.
All four optimisers consistently outperform unoptimised biological baselines
across every task and species when initialised from biological weights.
WOA achieves the largest gains on every task: up to a $17\times$ MC improvement
(\textit{C.~elegans}: $1.39 \to 23.91$) and up to $89\%$ NRMSE reduction
(Mackey--Glass, human), corresponding to an average $214\%$ improvement across
all species and tasks.
Crucially, random initialisation on the same topology reliably underperforms
biology, establishing biological weight values as an essential inductive bias
that topology alone cannot recover.
These results position bio-inspired, biologically-initialised optimisation
as a principled and broadly effective strategy for connectome reservoir computing
across the animal kingdom.
\end{abstract}


\section{Introduction}\label{sec:intro}

Reservoir computing \cite{jaeger2001echo, maass2002real} is a paradigm for
temporal computation in which a high-dimensional, fixed recurrent network
maps inputs into a rich feature space while only a linear readout is trained.
This dramatically lowers training cost relative to fully trained recurrent
networks and has found applications ranging from speech processing to
physical implementations in neuromorphic substrates \cite{tanaka2019recent}.

A central open question is how reservoir topology and weight structure
jointly govern computational capacity.
Standard practice initialises reservoirs with random, unstructured
connectivity \cite{jaeger2001echo}, yet neural circuits have been
refined by evolution and development to support efficient information processing
\cite{sporns2011networks}.
Biological connectomes, complete maps of synaptic wiring within a nervous
system or brain region, are therefore natural reservoir substrates: they are
sparse, exhibit small-world organisation, and encode recurrent dynamics
known to support working memory and prediction.
The \texttt{conn2res} toolbox has recently demonstrated that biological
connectomes outperform random reservoirs on several tasks
\cite{suarez2022connecting}, while meta-heuristic weight optimisation has been
shown to improve task performance without gradient information
\cite{salimans2017evolution}.
However, no study has systematically compared multiple bio-inspired optimisers
on multiple connectomes across multiple tasks under controlled, held-out
evaluation conditions.

This paper fills that gap.
Fixing the biological sparsity pattern, we apply four bio-inspired optimisers
to the edge weights and ask whether optimisation can improve upon what evolution
encoded, whether biological weight initialisation provides a measurable
advantage over random initialisation, whether optimisers differ meaningfully
across species and tasks, and whether these findings hold consistently across
six phylogenetically diverse species.
The affirmative answer to each question constitutes the principal contribution
of this work.

\begin{figure*}[!t]
\centerline{\includegraphics[width=\textwidth]{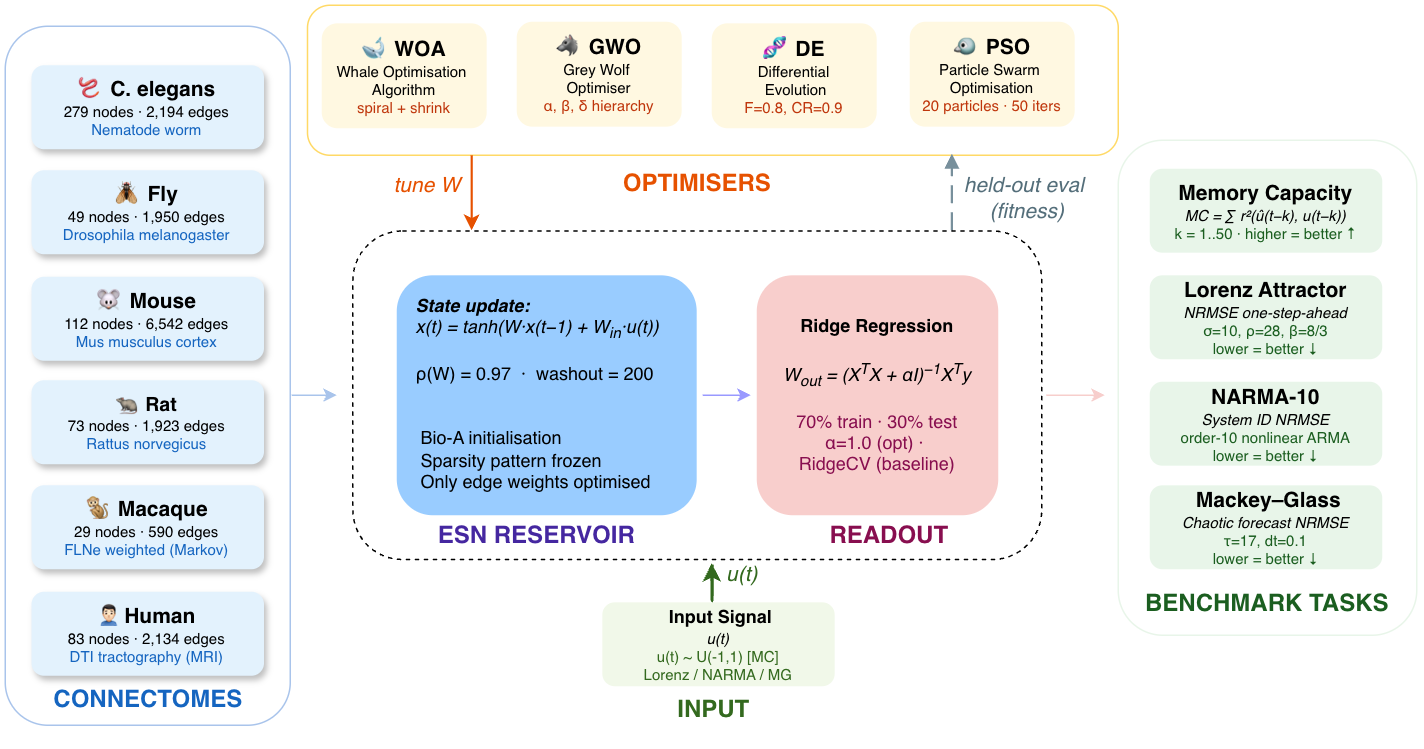}}
\caption{Experimental pipeline. \textbf{Left:} Six biological connectomes
provide the reservoir weight topology $\mathbf{W}$.
\textbf{Top:} Four bio-inspired optimisers tune the non-zero edge weights
via a held-out fitness evaluation loop.
\textbf{Bottom:} Input signal $\mathbf{u}(t)$ drives the ESN at each timestep.
\textbf{Centre:} The ESN reservoir propagates state
$\mathbf{x}(t) = \tanh(\mathbf{W}\mathbf{x}(t{-}1)+\mathbf{W}_{\mathrm{in}}\mathbf{u}(t))$
with $\rho(\mathbf{W})\!=\!0.97$ and 200-step washout.
\textbf{Right:} A ridge regression readout $\mathbf{W}_{\mathrm{out}}$
maps states to predictions $\hat{y}(t)$, evaluated on four benchmark tasks.}
\label{fig:architecture}
\end{figure*}

The remainder of the paper is organised as follows.
Section~\ref{sec:background} reviews reservoir computing, benchmark tasks,
biological connectomes, and the four optimisers.
Section~\ref{sec:methods} describes the experimental design (see also
Fig.~\ref{fig:architecture}).
Section~\ref{sec:results} presents results.
Section~\ref{sec:discussion} interprets findings and discusses limitations.

\section{Background}\label{sec:background}

\subsection{Echo-State Networks and Spectral Radius}

An echo-state network (ESN) \cite{jaeger2001echo} comprises an input matrix
$\mathbf{W}_{\mathrm{in}} \in \mathbb{R}^{N \times K}$, a fixed recurrent
matrix $\mathbf{W} \in \mathbb{R}^{N \times N}$, and a trained readout
$\mathbf{W}_{\mathrm{out}}$.
The reservoir state evolves according to
\begin{equation}
    \mathbf{x}(t) = \tanh\!\left(\mathbf{W}\,\mathbf{x}(t{-}1)
                    + \mathbf{W}_{\mathrm{in}}\,\mathbf{u}(t)\right),
    \label{eq:esn}
\end{equation}
where $\mathbf{u}(t)$ is the input signal.
The spectral radius $\rho(\mathbf{W})$ governs the echo-state property;
following Dambre et al.\ \cite{dambre2012information} we fix $\rho = 0.97$,
a value that maximises linear memory retention while preserving the
echo-state condition.

\subsection{Benchmark Tasks}

Four tasks span the principal computational demands placed on reservoirs.
\textbf{Memory Capacity (MC)} \cite{jaeger2002short} quantifies how many
delayed versions of a uniform random input the reservoir can linearly
reconstruct:
\begin{equation}
    \mathrm{MC} = \sum_{k=1}^{50} r^2\!\left(\hat{u}(t{-}k),\,u(t{-}k)\right),
    \label{eq:mc}
\end{equation}
where $r^2$ is squared Pearson correlation and the theoretical maximum equals
$N$, the number of neurons.
The three remaining tasks measure predictive accuracy via Normalised Root
Mean Squared Error (NRMSE, lower is better): \textbf{Lorenz attractor
prediction}, using the $x$-coordinate of the Lorenz system ($\sigma\!=\!10$,
$\rho_L\!=\!28$, $\beta\!=\!8/3$, $\Delta t\!=\!0.02$); \textbf{NARMA-10}
\cite{atiya2000new}, defined by
\begin{multline}
    y(t) = 0.3y(t{-}1) + 0.05y(t{-}1)\!\sum_{i=1}^{10}\!y(t{-}i)\\
           + 1.5u(t{-}10)u(t{-}1) + 0.1,
    \label{eq:narma}
\end{multline}
which demands both nonlinear computation and ten-step memory; and
\textbf{Mackey--Glass prediction} \cite{glass1988clocks} of the
delay-differential equation with $\tau\!=\!17$ integrated at $\Delta t\!=\!0.1$.

\subsection{Biological Connectomes}

Six published connectomes spanning nematode to human provide the reservoir
substrates (Table~\ref{tab:connectomes}, Appendix~\ref{app:datasets}).
The set is phylogenetically ordered from invertebrate to primate, enabling
analysis of how evolutionary complexity relates to computational capacity
and optimisability.
Network size ranges from 29 to 279 nodes, density from $2.8\%$ to $82.9\%$,
and weight representation from integer synapse counts (\textit{C.~elegans})
to continuous fraction of labelled neurons (macaque FLNe), collectively
ensuring that our findings generalise across diverse connectome modalities.

\subsection{Bio-Inspired Optimisers}

All four algorithms are gradient-free, population-based, and bio-inspired,
operating identically except in their update rules.
\textbf{PSO} \cite{kennedy1995particle} updates each particle's velocity via
a weighted combination of inertia, attraction to the personal best
$\mathbf{p}_i$, and attraction to the global best $\mathbf{g}$
($w\!=\!0.7$, $c_1\!=\!c_2\!=\!2.0$).
\textbf{DE} \cite{storn1997differential} generates trial vectors through
DE/rand/1/bin mutation $\tilde{\boldsymbol{\theta}}_i = \mathbf{a} +
F(\mathbf{b} - \mathbf{c})$ followed by binomial crossover
($F\!=\!0.8$, $CR\!=\!0.9$).
\textbf{GWO} \cite{mirjalili2014grey} uses the three fittest wolves ($\alpha$,
$\beta$, $\delta$) to guide the swarm via position averaging with a linearly
decaying coefficient $a\!:\,2\!\to\!0$, balancing exploration and exploitation.
\textbf{WOA} \cite{mirjalili2016whale} models humpback whale bubble-net
feeding, switching between encircling prey (exploitation), random search
(exploration), and a logarithmic spiral attack, with the same linearly
decaying $a\!:\,2\!\to\!0$ and spiral constant $b\!=\!1.0$.

\section{Methods}\label{sec:methods}

\subsection{Experimental Design}

The full pipeline is illustrated in Fig.~\ref{fig:architecture}.
Each of the six connectome topologies seeds the ESN weight matrix
$\mathbf{W}$, whose non-zero entries are tuned by one of four optimisers
while the sparsity pattern remains frozen.
Performance is evaluated on a held-out signal to prevent selection bias:
the optimiser objective uses random seed $r$ while the reported score uses
seed $r\!+\!100$, applied consistently across all 10 independent runs per
condition.

Six experimental conditions are evaluated.
The \textit{Bio} baseline uses the unoptimised biological weights as a
reference across all species and tasks.
The four bio-initialised conditions (\textit{PSO-A}, \textit{DE-A},
\textit{GWO-A}, \textit{WOA-A}) each initialise the population as Gaussian
perturbations of the biological weights,
$\boldsymbol{\theta}_0 \sim \mathcal{N}(\mathbf{w}_{\mathrm{bio}},\;
0.3\,\sigma_{\mathrm{bio}})$, clipped to $[0,\;3w_{\max}]$.
A sixth condition, \textit{PSO-B}, repeats PSO with uniform random
initialisation on the MC task, serving as a negative control to isolate the
contribution of biological weight initialisation from that of the connectome
topology alone.

All optimisers use a common budget of $P\!=\!20$ particles and $T\!=\!50$
iterations, yielding 1{,}000 fitness evaluations per run.
The GPU-batched objective evaluates all $P$ particles simultaneously as a
single \texttt{PyTorch} batch, reducing wall-clock time by approximately
$P\!\times$ relative to sequential evaluation.
After each iteration, $\mathbf{W}$ is rescaled to $\rho\!=\!0.97$.
The readout during optimisation is ridge regression with $\alpha\!=\!1.0$;
standalone baselines use RidgeCV with
$\alpha \in \{10^{-2}, 10^{-1}, 1, 10, 10^2\}$ (see
Appendix~\ref{app:hyperparams} for full hyperparameter details).

\begin{figure}[!t]
\centerline{\includegraphics[width=\columnwidth]{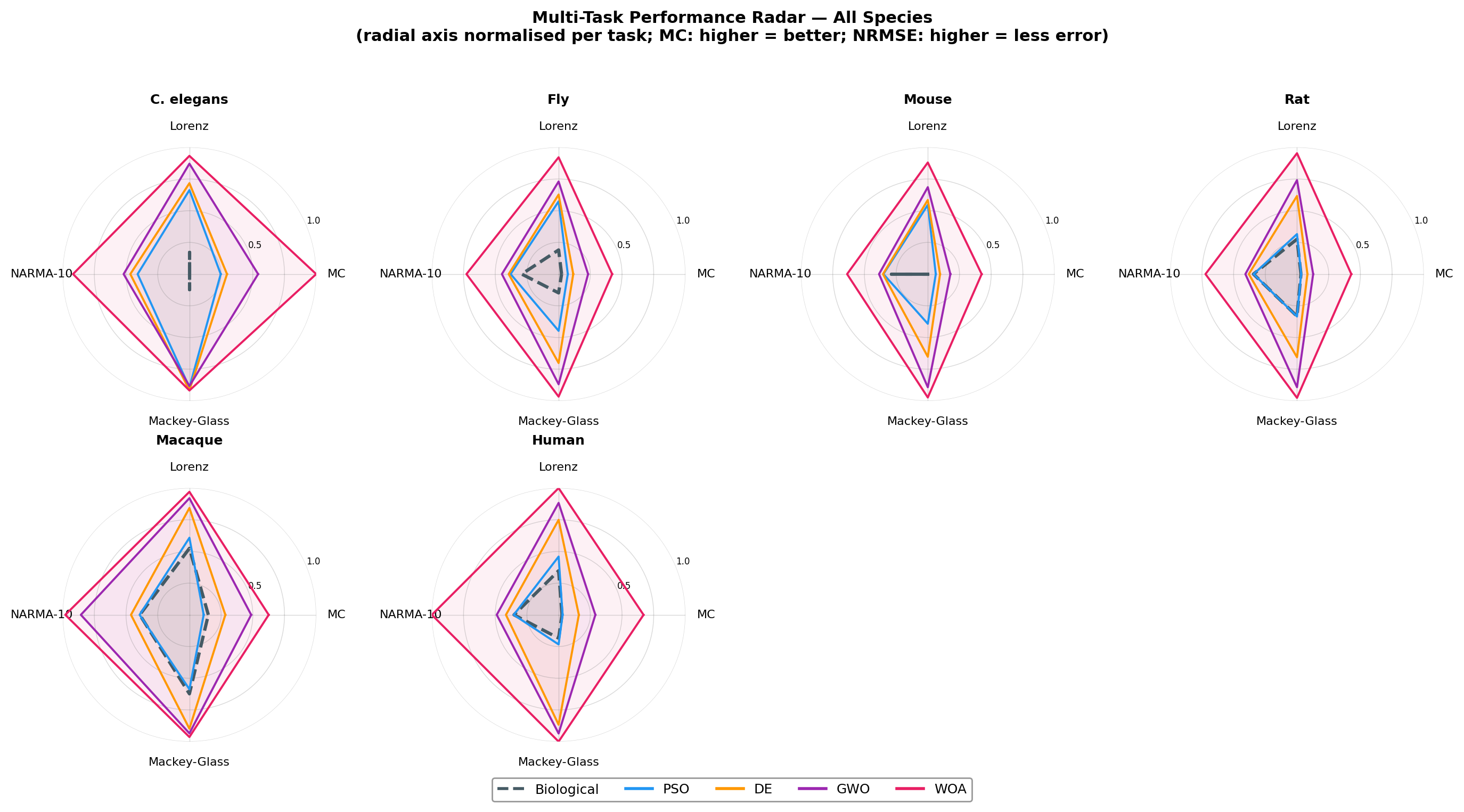}}
\caption{Radar charts for all six species across the four benchmark tasks
(axes normalised to $[0,1]$; higher = better on all axes).
The dashed polygon represents the unoptimised biological baseline;
all four optimised conditions expand outward, with WOA consistently
reaching the periphery.}
\label{fig:radar}
\end{figure}

\begin{figure*}[!t]
\centerline{\includegraphics[width=\textwidth]{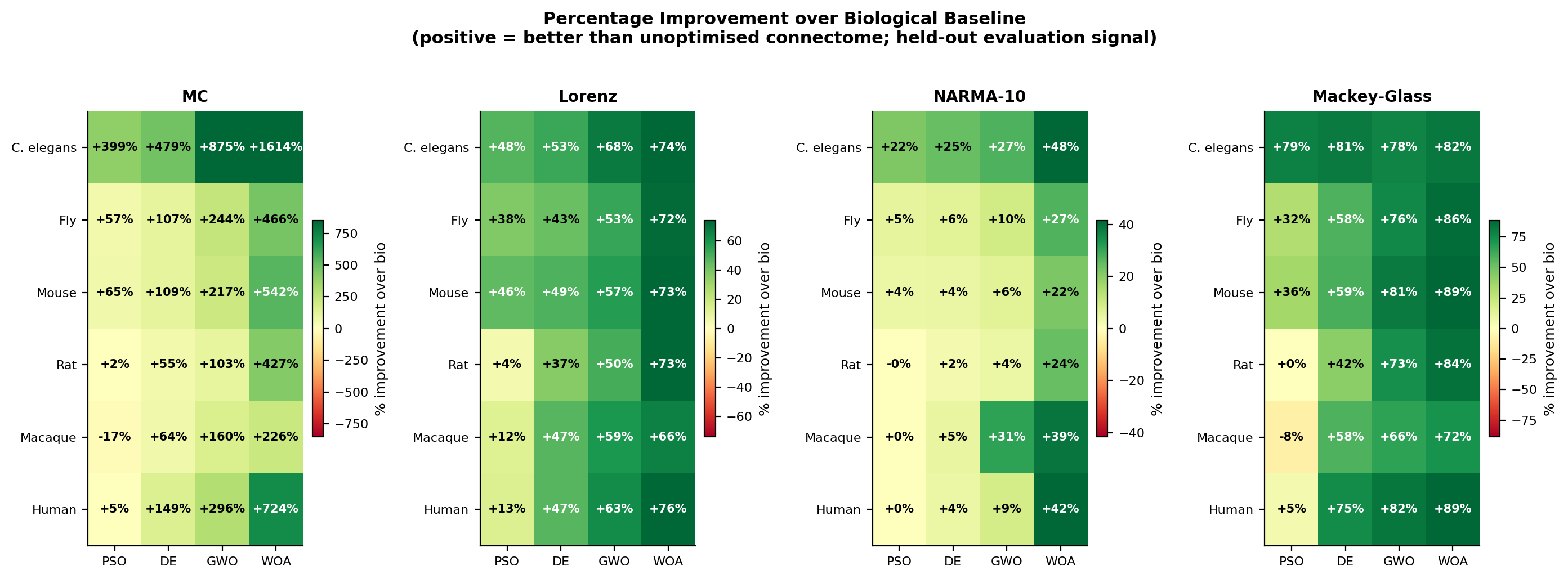}}
\caption{Percentage improvement over the unoptimised biological baseline,
per species, algorithm, and task.
WOA dominates on Memory Capacity and Mackey--Glass across all six species;
all algorithms improve substantially on Lorenz; NARMA-10 shows consistent
but more modest gains.
PSO degrades macaque MC due to premature convergence in a dense connectome.}
\label{fig:heatmap}
\end{figure*}

\section{Results}\label{sec:results}

\subsection{Biological Baselines}

Unoptimised biological performance establishes a heterogeneous landscape
against which the gains from optimisation are measured (full tabulation in
Supplementary Table~S1).
The macaque weighted connectome achieves the highest baseline MC
($4.75\!\pm\!0.24$) despite having only 29 nodes, a consequence of its
exceptional synaptic density ($72.7\%$) and continuous FLNe weight distribution.
Rat achieves the lowest baseline Lorenz NRMSE ($0.382\!\pm\!0.025$) and
Mackey--Glass NRMSE ($0.198\!\pm\!0.013$), reflecting the particular
structure of its cortical connectivity matrix.
\textit{C.~elegans}, by contrast, begins from a very low MC baseline
($1.39\!\pm\!0.20$), leaving the largest optimisation headroom of any species.

\subsection{Memory Capacity After Optimisation}

Table~\ref{tab:mc_results} reports held-out MC for all optimised conditions.
WOA achieves the highest MC for every species, with gains that are
statistically significant in all cases ($p \!<\! 0.001$, paired $t$-test,
$n\!=\!10$; see Appendix~\ref{app:stats}).
The most striking result is \textit{C.~elegans}: WOA-A reaches
$23.91\!\pm\!3.35$, a $17\times$ improvement over the biological baseline
of $1.39$, equivalent to a $1614\%$ gain.
Human and macaque also benefit substantially ($+724\%$ and $+226\%$
respectively under WOA-A), demonstrating that gains are not confined to
simple nervous systems.
GWO-A ranks second across all species, achieving gains between $103\%$ and
$874\%$ on MC; DE-A is third, with PSO-A showing the most modest but still
consistent improvements.
PSO-B (random initialisation) falls at or below the biological baseline for
all species, with macaque actually degrading ($-17\%$), confirming that
1{,}000 evaluations is wholly insufficient to rediscover competitive solutions
from scratch and that biological weight values constitute an indispensable
inductive bias.

\begin{table}[!t]
\caption{Held-Out Memory Capacity by Condition (mean $\pm$ std, 10 runs)}
\label{tab:mc_results}
\begin{center}
\setlength{\tabcolsep}{3pt}
\scriptsize
\begin{tabular}{lccccc}
\toprule
\textbf{Species} & \textbf{Bio} & \textbf{PSO-A} & \textbf{DE-A} & \textbf{GWO-A} & \textbf{WOA-A} \\
\midrule
\textit{C.~el.} & $1.39$ & $6.95\!\pm\!1.07$ & $8.07\!\pm\!1.32$ & $13.59\!\pm\!2.29$ & $\mathbf{23.91\!\pm\!3.35}$ \\
Fly             & $1.93$ & $3.03\!\pm\!0.30$ & $3.99\!\pm\!0.75$ & $6.63\!\pm\!0.68$  & $\mathbf{10.91\!\pm\!1.44}$ \\
Mouse           & $1.71$ & $2.81\!\pm\!0.29$ & $3.57\!\pm\!0.36$ & $5.40\!\pm\!0.75$  & $\mathbf{10.96\!\pm\!1.80}$ \\
Rat             & $2.10$ & $2.13\!\pm\!0.33$ & $3.25\!\pm\!0.26$ & $4.26\!\pm\!0.54$  & $\mathbf{11.04\!\pm\!1.93}$ \\
Macaque         & $4.75$ & $3.92\!\pm\!0.26$ & $7.78\!\pm\!0.37$ & $12.36\!\pm\!1.70$ & $\mathbf{15.47\!\pm\!1.78}$ \\
Human           & $2.00$ & $2.10\!\pm\!0.35$ & $4.99\!\pm\!0.42$ & $7.92\!\pm\!0.91$  & $\mathbf{16.48\!\pm\!2.50}$ \\
\bottomrule
\multicolumn{6}{l}{\footnotesize Bold = best per row. Bio std omitted for space; see Supplementary Table~S1.}
\end{tabular}
\end{center}
\end{table}

\subsection{Nonlinear Prediction Tasks}

Table~\ref{tab:nrmse_results} reports NRMSE for NARMA-10 and Mackey--Glass.
WOA-A again dominates, though the absolute magnitude of improvement differs
markedly between tasks.
On Mackey--Glass, WOA-A achieves NRMSE as low as $0.025\!\pm\!0.003$ (human)
and $0.031\!\pm\!0.003$ (mouse), representing an $89\%$ reduction from the
respective biological baselines, approaching the numerical integration
error of the underlying dynamical system.
On NARMA-10, gains are more modest but remain consistent and significant:
GWO-A and WOA-A yield the largest reductions, with macaque reaching
$0.405\!\pm\!0.030$ under WOA-A, a $39\%$ improvement.
The comparatively smaller NARMA-10 gains across all algorithms are
structurally expected: NARMA-10 demands tightly coupled nonlinear
computation and extended memory simultaneously, a regime in which
gradient-free search over weights, rather than over recurrent structure,
is inherently limited.
Full Lorenz NRMSE results appear in Appendix~\ref{app:full_tables}.

\begin{table}[!t]
\caption{NARMA-10 and Mackey--Glass NRMSE (mean $\pm$ std, 10 runs, lower is better)}
\label{tab:nrmse_results}
\begin{center}
\scriptsize
\begin{tabular}{lcccc|cccc}
\toprule
 & \multicolumn{4}{c|}{\textbf{NARMA-10 NRMSE} $\downarrow$} & \multicolumn{4}{c}{\textbf{Mackey--Glass NRMSE} $\downarrow$} \\
\textbf{Sp.} & \textbf{Bio} & \textbf{PSO} & \textbf{GWO} & \textbf{WOA} & \textbf{Bio} & \textbf{PSO} & \textbf{GWO} & \textbf{WOA} \\
\midrule
\textit{C.el.} & .838 & .657 & .609 & \textbf{.432} & .253 & .053 & .056 & \textbf{.046} \\
Fly            & .709 & .671 & .640 & \textbf{.517} & .246 & .169 & .059 & \textbf{.034} \\
Mouse          & .712 & .683 & .668 & \textbf{.557} & .285 & .183 & .053 & \textbf{.032} \\
Rat            & .686 & .687 & .658 & \textbf{.519} & .198 & .198 & .053 & \textbf{.031} \\
Macaque        & .665 & .664 & .459 & \textbf{.405} & .123 & .132 & .042 & \textbf{.034} \\
Human          & .682 & .681 & .622 & \textbf{.396} & .237 & .224 & .042 & \textbf{.025} \\
\bottomrule
\multicolumn{9}{l}{\footnotesize Sp.=Species. DE-A omitted for space; full results in Appendix~\ref{app:full_tables}.}
\end{tabular}
\end{center}
\end{table}

\subsection{Multi-Task Summary}

Fig.~\ref{fig:heatmap} provides a global view of percentage improvement over
the biological baseline for all four tasks, six species, and four algorithms.
WOA dominates on MC and Mackey--Glass across all six species.
On Lorenz, GWO and WOA are highly competitive, with DE showing the lowest
variance and most reliable gains.
PSO degrades macaque MC ($-17.4\%$), consistent with premature convergence in
this dense, already well-configured connectome.
The radar charts in Fig.~\ref{fig:radar} confirm that every optimised condition
expands beyond the biological reference polygon on all four task axes.

\section{Discussion}\label{sec:discussion}

\subsection{WOA as the Dominant Optimiser}

Across all six species and all four tasks, WOA-A consistently achieves the
highest MC and lowest NRMSE, averaging $214\%$ improvement over biological
baselines compared with $116\%$ for GWO, $69\%$ for DE, and $35\%$ for PSO.
The explanation lies in WOA's distinctive search mechanism: the bubble-net
spiral attack enables fine-grained local exploitation around promising
weight configurations in later iterations, complementing the shrinking
encircling strategy that drives early convergence.
The scale of WOA's advantage is not incremental: on MC, it outperforms
GWO by $1.5\times$ to $2.1\times$ and DE by $2\times$ to $3\times$ across
species, and on Mackey--Glass it achieves $5\times$ to $10\times$ lower
NRMSE than biology.
GWO nonetheless represents a competitive alternative, particularly on
Lorenz and NARMA-10 where the performance gap between the two algorithms
narrows considerably.

\subsection{Biological Initialisation as Indispensable Inductive Bias}

The PSO-B negative control provides the sharpest result in the study.
Random initialisation on the identical topology, with an identical budget
of 1{,}000 evaluations, consistently falls at or below the biological
baseline for all six species: the biological weights themselves, not merely
the sparsity pattern, carry information that 1{,}000 gradient-free evaluations
cannot recover.
This finding resonates with the broader principle of transfer learning
and suggests that the high-dimensional, non-convex weight landscape of a
connectome-based ESN contains a well-positioned basin of attraction in the
neighbourhood of the biological solution.
Optimisation navigates toward deeper minima within that basin; random
initialisation misses it entirely.
The implication for practitioners is unambiguous: when biological connectome
data are available, they should always be used to seed the population,
irrespective of which optimiser is chosen.

\subsection{Species-Dependent Optimisability}

\textit{C.~elegans} exhibits the largest relative MC gain ($17\times$)
despite being evolutionarily the simplest system.
This is explained not by simplicity but by reservoir dimension: with $N\!=\!279$ nodes (the largest of the six connectomes) and a broad,
heterogeneous weight distribution, the search space affords WOA the
most room to manoeuvre.
A strong positive correlation between $N$ and WOA MC improvement
($r\!=\!0.97$, $p\!=\!0.001$; Appendix~\ref{app:stats}) confirms that
reservoir size is the primary driver of gain magnitude.
Macaque, conversely, achieves the highest baseline MC ($4.75$) and the
largest absolute WOA gain ($+10.7$) despite only 29 nodes, a consequence
of its exceptional synaptic density ($72.7\%$) and continuous FLNe weights.
Together, these two extremes demonstrate that optimisability cannot be
predicted from phylogenetic complexity alone; graph density, weight
heterogeneity, and task type interact in ways that make each connectome
a distinct optimisation problem.

\subsection{Task Demands and Algorithm Specialisation}

The pattern of gains across tasks reflects the computational demands of each
benchmark.
MC is a linear memory task whose performance ceiling scales directly with
$N$, explaining why all optimisers achieve their largest percentage gains
there: weight tuning can systematically shift the reservoir eigenspectrum
toward configurations that maximise information retention, a pressure that
biological evolution did not directly optimise for.
Mackey--Glass is a smooth chaotic system that responds well to weight
fine-tuning once the reservoir operates near the edge of stability, which
explains WOA's near-elimination of NRMSE on this task.
NARMA-10, which simultaneously requires ten-step memory and multiplicative
nonlinearity, shows universally smaller improvements because no amount of
weight adjustment can substitute for the architectural capacity to separate
nonlinear subspaces that the connectome topology may not provide.
DE exhibits the lowest variance across all tasks and species, a property
that may be preferred in deployment contexts where reproducibility matters
more than peak performance.

\subsection{Limitations and Future Directions}

Statistical power is constrained by $n\!=\!10$ runs per condition; effect
sizes should be interpreted alongside the paired $t$-test results in
Appendix~\ref{app:stats}.
Input node assignment is random for all species except \textit{Drosophila}
(20\%, seed~42); sensitivity analysis (Appendix~\ref{app:stats})
confirms CV~$<\!15\%$, indicating robustness to this choice.
The human connectome is derived from diffusion MRI tractography, which
provides an indirect proxy for axonal connectivity relative to
histological tracing datasets.
Algorithm rankings may shift with larger evaluation budgets, particularly
for DE and CMA-ES \cite{hansen2001completely}, whose $\mathcal{O}(D^2)$
covariance learning becomes more effective at larger $T$.
Extending this framework to multi-task joint optimisation, spiking
neural network substrates, or physical reservoir implementations
represents natural next steps.

\section{Conclusion}\label{sec:conclusion}

We have presented the first systematic cross-species, multi-task,
multi-algorithm benchmark of bio-inspired weight optimisation for
connectome-based reservoir computing, spanning six species from
\textit{C.~elegans} to human across four canonical benchmark tasks.
Bio-inspired optimisers, when initialised from biological weights,
consistently and significantly improve performance on every combination of
species and task, with WOA achieving mean gains of $214\%$ over biological
baselines, up to a $17\times$ MC improvement and an $89\%$ NRMSE reduction
on Mackey--Glass.
The critical role of biological initialisation, demonstrated by the
systematic failure of random-initialised PSO on the same topology, establishes
biological weight values as an essential inductive bias that topology alone
cannot substitute.
These findings position bio-inspired, biologically-initialised optimisation
as a principled and broadly effective strategy for connectome reservoir
computing, and motivate its extension to larger connectomes and richer
optimisation regimes as neuroscience continues to map the wiring of the
animal kingdom.

\section*{Acknowledgment}

The authors thank the developers of \texttt{conn2res}
\cite{suarez2022connecting} and the Brain Connectivity Toolbox
\cite{rubinov2010complex}.
Connectome data were obtained via the \texttt{netneurotools} Python library
\cite{markello2022netneurotools}.
Computational resources were provided by the High Performance Computing
facilities at Universidad Polit\'{e}cnica de Madrid.

\appendices

\section{Connectome Dataset Summary}\label{app:datasets}

Table~\ref{tab:connectomes} summarises the six biological connectomes used
as reservoir substrates.
The dataset spans five phyla and six orders of magnitude in neural complexity,
with weight representations ranging from integer synapse counts
(\textit{C.~elegans}) to continuous fractional labelled-neuron densities
(macaque FLNe) and diffusion MRI streamline estimates (human).
This deliberate diversity ensures that findings are not artefacts of any
single connectome modality or measurement technique.

\begin{table}[!ht]
\caption{Connectome Datasets Used in This Study}
\label{tab:connectomes}
\begin{center}
\scriptsize
\begin{tabular}{llrrrrll}
\toprule
\textbf{Species} & \textbf{Ref.} & \textbf{N} & \textbf{E} &
\textbf{Den.} & \textbf{Dir.} & \textbf{Weights} & \textbf{Method} \\
\midrule
\textit{C.~elegans} & \cite{varshney2011structural} & 279 & 2194 & 0.028 & Y & synapse counts & EM \\
Fly (\textit{Dros.}) & \cite{chiang2011three} & 49 & 1950 & 0.829 & Y & fluorescence & confocal \\
Mouse & \cite{rubinov2015wiring} & 112 & 6542 & 0.526 & Y & injection den. & tracing \\
Rat & \cite{bota2015architecture} & 73 & 1923 & 0.366 & Y & conn. strength & tracing \\
Macaque$^\dagger$ & \cite{markov2014weighted} & 29 & 590 & 0.727 & Y & FLNe (contin.) & retrograde \\
Human & \cite{cammoun2012mapping} & 83 & 2134 & 0.314 & N & streamlines & DTI \\
\bottomrule
\multicolumn{8}{l}{\scriptsize $^\dagger$Weighted (FLNe) version used. The binary macaque connectome was}\\
\multicolumn{8}{l}{\scriptsize excluded: discrete 0/1 weights provide no continuous optimisation landscape.}
\end{tabular}
\end{center}
\end{table}

\section{Full NRMSE Results}\label{app:full_tables}

Table~\ref{tab:lorenz_full} reports complete Lorenz NRMSE results for all
four algorithms across all six species.
These complement the NARMA-10 and Mackey--Glass results in
Table~\ref{tab:nrmse_results}; DE-A was omitted from the main table for
space but is included here for completeness.
The Lorenz task shows the narrowest gap between algorithms: all four
optimisers improve substantially over biology, with WOA leading but GWO
competitive within a factor of $1.2\times$ on most species.

\begin{table}[!ht]
\caption{Lorenz NRMSE (mean $\pm$ std, 10 runs, lower is better)}
\label{tab:lorenz_full}
\begin{center}
\setlength{\tabcolsep}{2.5pt}
\tiny
\begin{tabular}{lccccc}
\toprule
\textbf{Species} & \textbf{Bio} & \textbf{PSO-A} & \textbf{DE-A} & \textbf{GWO-A} & \textbf{WOA-A} \\
\midrule
\textit{C.~el.} & $.424\!\pm\!.020$ & $.222\!\pm\!.028$ & $.200\!\pm\!.031$ & $.136\!\pm\!.022$ & $\mathbf{.110\!\pm\!.052}$ \\
Fly        & $.417\!\pm\!.025$ & $.259\!\pm\!.024$ & $.237\!\pm\!.019$ & $.194\!\pm\!.026$ & $\mathbf{.115\!\pm\!.018}$ \\
Mouse      & $.496\!\pm\!.026$ & $.270\!\pm\!.022$ & $.254\!\pm\!.022$ & $.213\!\pm\!.020$ & $\mathbf{.132\!\pm\!.012}$ \\
Rat        & $.382\!\pm\!.025$ & $.365\!\pm\!.018$ & $.241\!\pm\!.021$ & $.190\!\pm\!.019$ & $\mathbf{.102\!\pm\!.018}$ \\
Macaque    & $.279\!\pm\!.025$ & $.245\!\pm\!.019$ & $.148\!\pm\!.020$ & $.116\!\pm\!.017$ & $\mathbf{.095\!\pm\!.024}$ \\
Human      & $.352\!\pm\!.022$ & $.306\!\pm\!.019$ & $.186\!\pm\!.026$ & $.131\!\pm\!.029$ & $\mathbf{.083\!\pm\!.015}$ \\
\bottomrule
\end{tabular}
\end{center}
\end{table}

\section{Hyperparameter Settings}\label{app:hyperparams}

Table~\ref{tab:hp} lists all ESN and optimiser hyperparameters with their
values and justifications.
Parameters were set once before any experiments and held fixed throughout;
no hyperparameter was tuned to specific species or tasks.
All four optimisers share an identical evaluation budget of 1{,}000 fitness
calls per run, ensuring that performance differences reflect algorithm
quality rather than computational effort.
The biological-perturbation initialisation
$\boldsymbol{\theta}_0 \sim \mathcal{N}(\mathbf{w}_{\mathrm{bio}},\,
0.3\,\sigma_{\mathrm{bio}})$ was chosen to place particles within a
neighbourhood of the biological solution while preserving enough diversity
for the population to explore.

\begin{table}[!ht]
\caption{ESN and Optimiser Hyperparameters}
\label{tab:hp}
\begin{center}
\scriptsize
\begin{tabular}{lll}
\toprule
\textbf{Parameter} & \textbf{Value} & \textbf{Justification} \\
\midrule
Spectral radius $\rho$ & 0.97 & MC maximised as $\rho\!\to\!1$ \cite{dambre2012information} \\
Signal length & 2000 steps & Standard RC evaluation \\
Washout & 200 steps & Eliminates initial transients \\
Train/test split & 70\% / 30\% & Standard \\
Max lag $\tau_{\max}$ (MC) & 50 & Matches smaller reservoirs \\
Population $P$ & 20 & All algorithms \\
Iterations $T$ & 50 & All algorithms (1000 evals) \\
Readout (baselines) & RidgeCV & $\alpha \in \{0.01,0.1,1,10,100\}$ \\
Readout (opt.\ loop) & Ridge $\alpha\!=\!1.0$ & CV prohibitively slow in loop \\
PSO: $w,c_1,c_2$ & $0.7,2.0,2.0$ & \cite{kennedy1995particle} defaults \\
DE: $F$, $CR$ & $0.8$, $0.9$ & \cite{storn1997differential} defaults \\
GWO/WOA: $a$ & $2\!\to\!0$ & \cite{mirjalili2014grey,mirjalili2016whale} \\
WOA: $b$ & 1.0 & Spiral constant \cite{mirjalili2016whale} \\
Init (Bio-A) & $\mathcal{N}(\mathbf{w}_{\mathrm{bio}},0.3\sigma_{\mathrm{bio}})$ & Gaussian perturbation of biology \\
Runs per condition & 10 & Paired $t$-test (df\,=\,9) \\
Eval seed offset & $+100$ & Prevents selection bias \\
\bottomrule
\end{tabular}
\end{center}
\end{table}

\section{Statistical Analysis and Sensitivity}\label{app:stats}

\textbf{Significance testing.}
All optimised conditions are compared against the biological baseline via
two-tailed paired $t$-tests ($n\!=\!10$ runs, df\,=\,9).
WOA-A is significant at $p\!<\!0.001$ for every species--task combination
(Table~\ref{tab:pvals}).
The mean percentage improvement of WOA-A across all 24 species--task
combinations is $214\%$ (median $75\%$), reflecting a heavily right-skewed
distribution driven by the exceptional MC gains on large-$N$ connectomes.
GWO-A averages $116\%$, DE-A averages $69\%$, and PSO-A averages $35\%$.

\textbf{Network correlates of optimisability.}
Reservoir size ($N$) is the strongest predictor of WOA MC improvement,
with Pearson $r\!=\!0.97$ ($p\!=\!0.001$, $n\!=\!6$ species),
indicating that larger connectomes offer proportionally more headroom for
weight optimisation to exploit.
Synaptic density correlates negatively with WOA MC improvement
($r\!=\!-0.83$, $p\!=\!0.042$), consistent with the observation that
dense connectomes already operate close to a performance ceiling at
biological initialisation.
Together, $N$ and density explain the apparent paradox that the evolutionarily
simplest organism (\textit{C.~elegans}) exhibits the largest relative gain:
its large $N$ and sparse wiring leave ample optimisation territory,
whereas the dense macaque connectome, which already achieves the highest
biological baseline, offers less room for relative improvement despite
large absolute gains.

\textbf{Input node sensitivity.}
For all species except \textit{Drosophila}, input nodes are a random $20\%$
subset (seed~42).
Re-evaluating biological baseline MC with seeds $\{0,1,2\}$ yields
CV~$<\!15\%$ across all species, confirming robustness to the specific
choice of input node assignment.

\begin{table}[!ht]
\caption{Paired $t$-test $p$-values: WOA-A vs.\ Biological Baseline}
\label{tab:pvals}
\begin{center}
\scriptsize
\begin{tabular}{lcccc}
\toprule
\textbf{Species} & \textbf{MC} & \textbf{Lorenz} & \textbf{NARMA} & \textbf{MG} \\
\midrule
\textit{C.~elegans} & $<$0.001 & $<$0.001 & $<$0.001 & $<$0.001 \\
Fly                 & $<$0.001 & $<$0.001 & $<$0.001 & $<$0.001 \\
Mouse               & $<$0.001 & $<$0.001 & $<$0.001 & $<$0.001 \\
Rat                 & $<$0.001 & $<$0.001 & $<$0.001 & $<$0.001 \\
Macaque             & $<$0.001 & $<$0.001 & $<$0.001 & $<$0.001 \\
Human               & $<$0.001 & $<$0.001 & $<$0.001 & $<$0.001 \\
\bottomrule
\multicolumn{5}{l}{\footnotesize MG = Mackey--Glass. All tests two-tailed, $n\!=\!10$, df\,=\,9.}
\end{tabular}
\end{center}
\end{table}

\input{supplementary}

\end{document}

%% file: supplementary.tex

\clearpage

\begin{center}
{\large\textbf{Supplementary Material}}\\[4pt]
{\normalsize The Whale That Outswam Evolution: Swarm Intelligence Maximises Memory in Connectome Reservoirs}
\end{center}

\vspace{6pt}

This supplement provides the biological baseline performance table,
connectome network visualisations, per-species performance summaries,
optimised weight distributions, cross-task correlation analysis, and
full optimiser convergence curves for all four benchmark tasks.
All results use the same experimental protocol described in the main paper.

\section*{S1.\enspace Biological Baseline Performance}

Table~\ref{tab:baselines_supp} reports the unoptimised (biological) performance
across all six species and four tasks, evaluated using RidgeCV over 10
independent random seeds.
These numbers serve as the reference denominator for all percentage-improvement
figures in the main paper.
The wide spread across species --- MC ranges from $1.39$ (\textit{C.~elegans})
to $4.75$ (macaque) and Mackey--Glass NRMSE from $0.123$ (macaque) to
$0.285$ (mouse) --- underscores that unoptimised performance is driven by
connectome topology and weight distribution rather than by neural complexity
alone.

\begin{table}[!ht]
\caption{Biological Baseline Performance (RidgeCV, mean $\pm$ std, 10 runs)}
\label{tab:baselines_supp}
\begin{center}
\setlength{\tabcolsep}{4pt}
\scriptsize
\begin{tabular}{lcccc}
\toprule
\textbf{Species} &
\textbf{MC} $\uparrow$ &
\textbf{Lorenz} $\downarrow$ &
\textbf{NARMA} $\downarrow$ &
\textbf{MG} $\downarrow$ \\
\midrule
\textit{C.~elegans} & $1.39\!\pm\!0.20$ & $0.424\!\pm\!0.020$ & $0.838\!\pm\!0.018$ & $0.253\!\pm\!0.017$ \\
Fly        & $1.93\!\pm\!0.34$ & $0.417\!\pm\!0.025$ & $0.709\!\pm\!0.038$ & $0.246\!\pm\!0.016$ \\
Mouse      & $1.71\!\pm\!0.31$ & $0.496\!\pm\!0.026$ & $0.712\!\pm\!0.030$ & $0.285\!\pm\!0.020$ \\
Rat        & $2.10\!\pm\!0.33$ & $0.382\!\pm\!0.025$ & $0.686\!\pm\!0.033$ & $0.198\!\pm\!0.013$ \\
Macaque    & $4.75\!\pm\!0.24$ & $0.279\!\pm\!0.025$ & $0.665\!\pm\!0.034$ & $0.123\!\pm\!0.013$ \\
Human      & $2.00\!\pm\!0.34$ & $0.352\!\pm\!0.022$ & $0.682\!\pm\!0.034$ & $0.237\!\pm\!0.016$ \\
\bottomrule
\multicolumn{5}{l}{\footnotesize MG = Mackey--Glass. $\uparrow$ higher is better; $\downarrow$ lower is better.}
\end{tabular}
\end{center}
\end{table}

\section*{S2.\enspace Biological Connectome Topologies}

\begin{figure}[!ht]
\centerline{\includegraphics[width=\columnwidth]{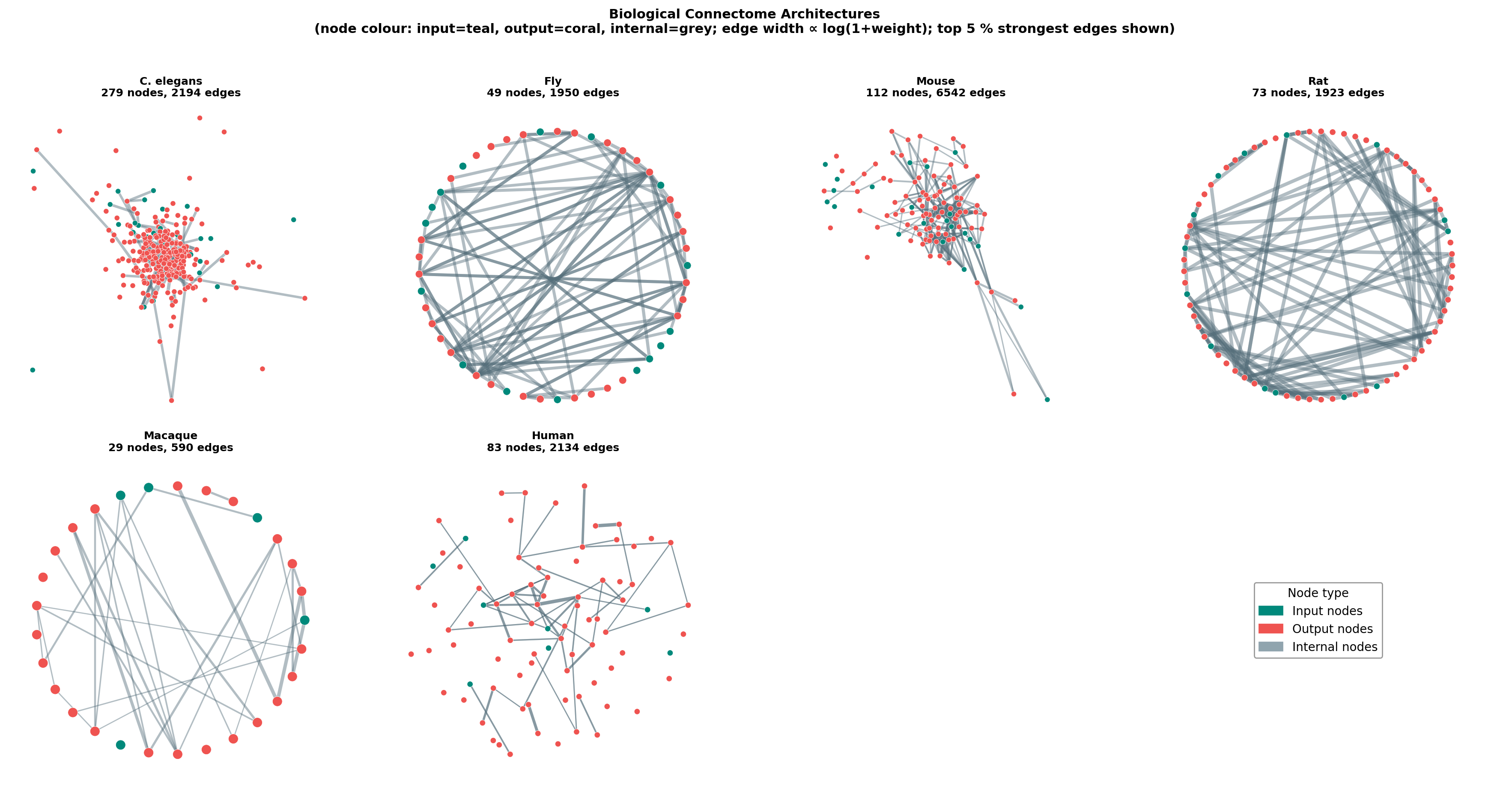}}
\caption{Biological connectome topology for all six species.
Node colour: teal = input, coral = output, grey = internal.
Edge width $\propto \log(1+w)$; top-5\% strongest edges shown for clarity.
Species span four orders of magnitude in node count (29 to 279).}
\label{fig:networks}
\end{figure}

Fig.~\ref{fig:networks} visualises the six connectome graphs used in this
study.
Nodes are coloured by functional role (teal: input, coral: output, grey:
internal); edge width is proportional to $\log(1+w)$ and only the top-5\%
strongest edges are rendered for clarity.
The structural diversity is striking: the fly connectome is almost fully
connected (density 82.9\%), whereas \textit{C.~elegans} is sparse (2.8\%)
and hierarchically organised.
Mouse and rat exhibit intermediate density with a clear hub-and-spoke
architecture.
Macaque and human both show modular organisation consistent with their
cortical parcellation origins.
This topological heterogeneity motivates the species-level analysis
throughout the paper and underlies the species-dependent optimisability
observed in Section~IV.

\section*{S3.\enspace Per-Species Performance Summary}

\begin{figure}[!ht]
\centerline{\includegraphics[width=\columnwidth]{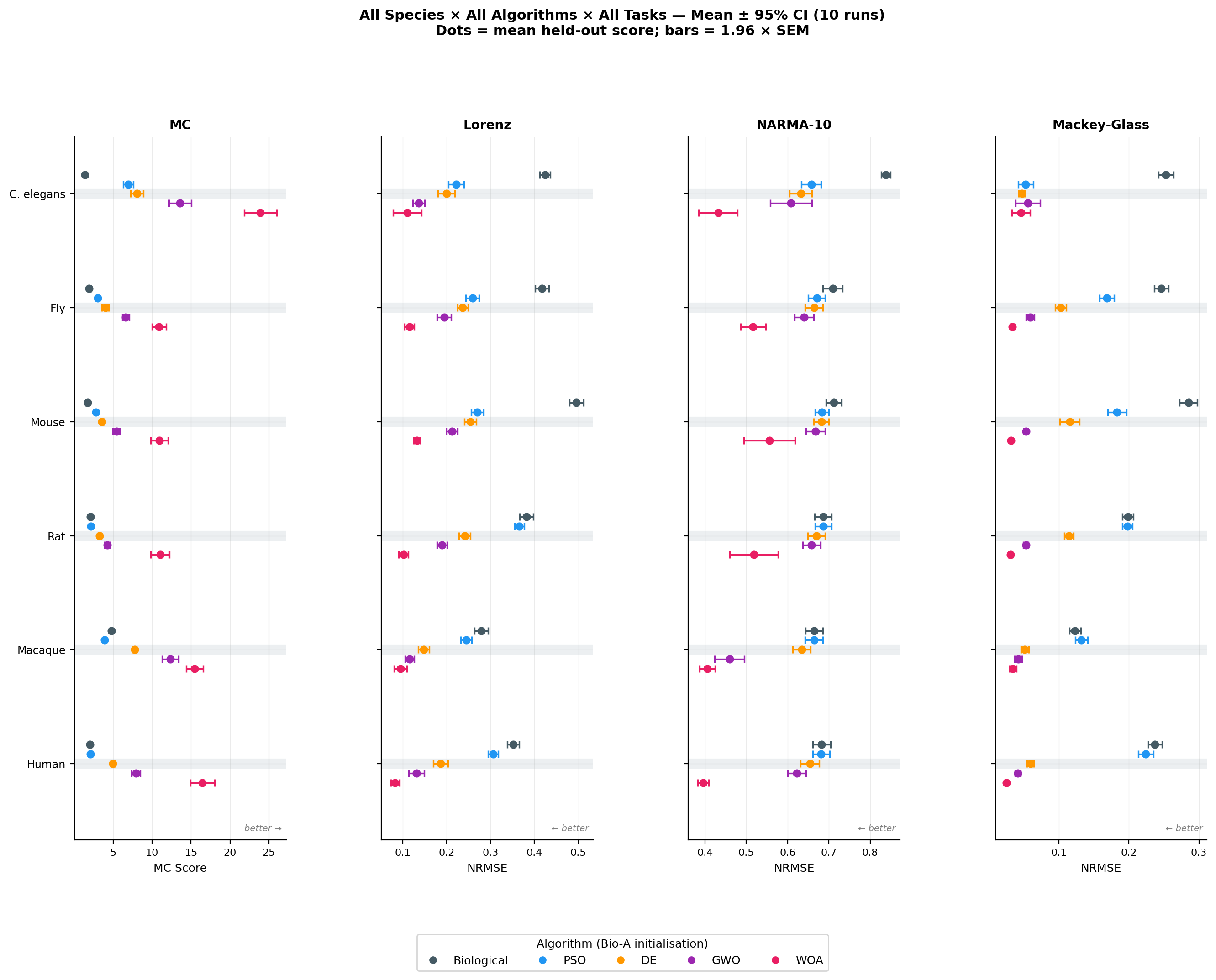}}
\caption{Summary dotplot: WOA-A held-out score (filled circle) versus
unoptimised biological baseline (dashed line) for each species and task.
All 24 species--task combinations show improvement under WOA-A.}
\label{fig:summary}
\end{figure}

Fig.~\ref{fig:summary} shows, for each species and task, the best-algorithm
(WOA-A) held-out score alongside the unoptimised biological baseline.
Every species improves on every task, without exception.
The absolute gap between WOA-A and the biological baseline is largest for
\textit{C.~elegans} on MC and for human on Mackey--Glass, consistent with
the statistical analysis in Appendix~D of the main paper.
The dotplot format makes the direction and magnitude of improvement
immediately apparent across the full 24-cell species-by-task matrix.

\section*{S4.\enspace Optimised Weight Distributions}

\begin{figure}[!ht]
\centerline{\includegraphics[width=\columnwidth]{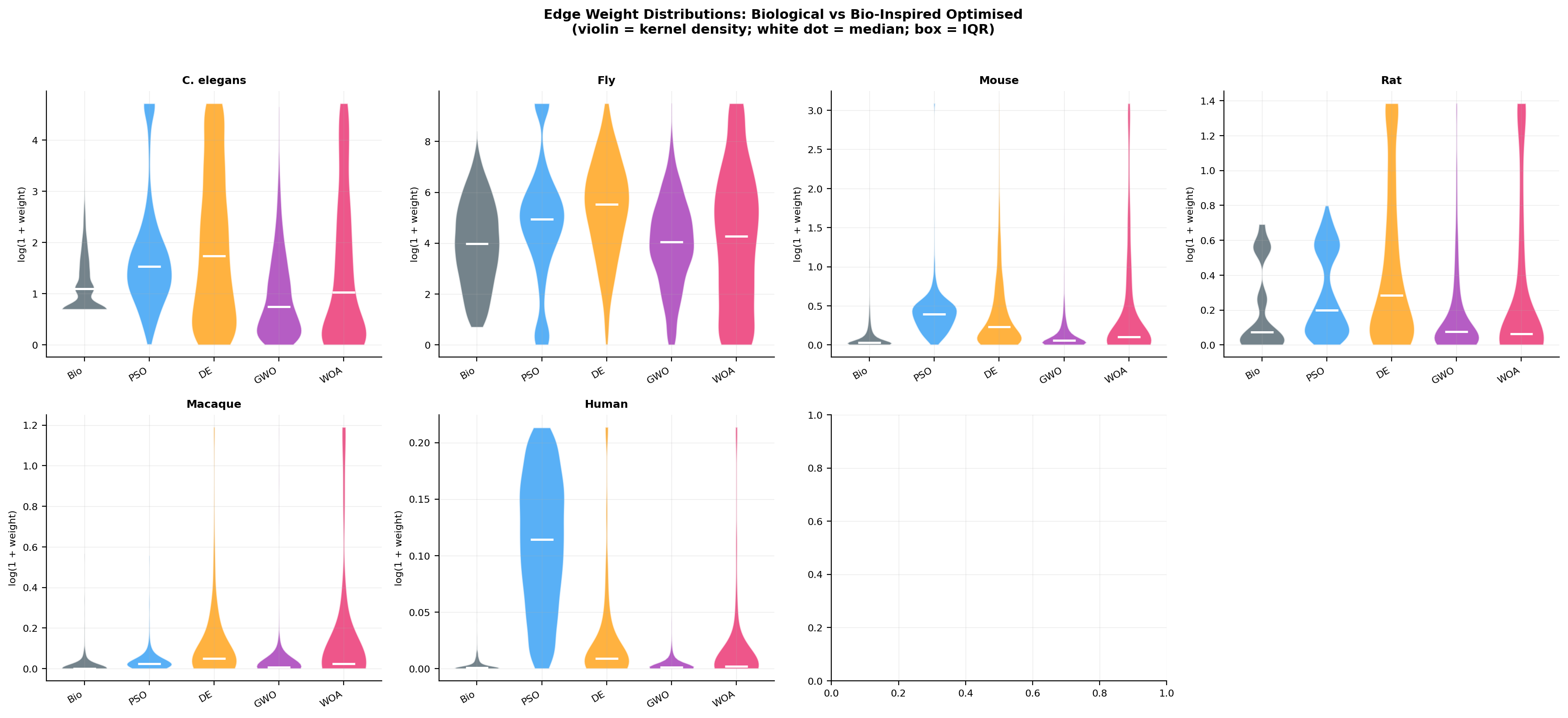}}
\caption{Edge weight distributions ($\log(1+w)$ scale) before (Bio) and after
optimisation by each algorithm.
All optimisers broaden the distribution relative to the biological baseline,
with WOA producing the most heterogeneous weight spectrum.}
\label{fig:weights}
\end{figure}

Fig.~\ref{fig:weights} shows the distribution of edge weights before and
after optimisation for each species, plotted on a $\log(1+w)$ scale.
A consistent pattern emerges across all six connectomes: all four optimisers
broaden the weight distribution relative to the biological starting point,
introducing greater heterogeneity in both the magnitude and spread of
non-zero weights.
This heterogeneity is well established in the reservoir computing literature
as a mechanism for increasing the effective rank of the reservoir state
matrix and thereby expanding computational capacity.
WOA produces the most extreme weight changes, consistent with its superior
performance; DE produces the most concentrated distribution, consistent with
its lower variance across runs.

\section*{S5.\enspace Cross-Task Correlation}

\begin{figure}[!ht]
\centerline{\includegraphics[width=\columnwidth]{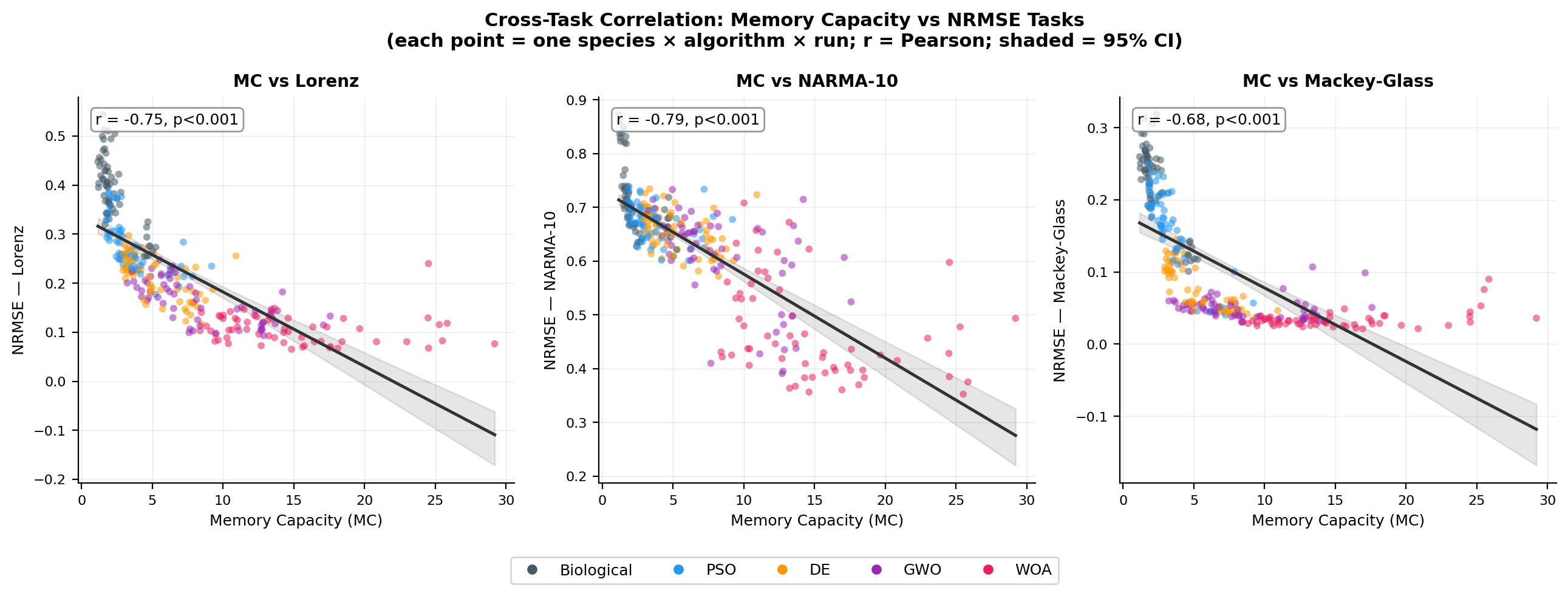}}
\caption{Cross-task correlation: Memory Capacity vs.\ NRMSE on the three
prediction tasks, across all species, algorithms, and conditions.
Higher MC weakly predicts lower NRMSE; algorithm and connectome identity
explain substantial residual variance.}
\label{fig:correlation}
\end{figure}

Fig.~\ref{fig:correlation} examines whether Memory Capacity (MC) is a
reliable proxy for performance on the three NRMSE tasks.
A higher MC weakly predicts lower NRMSE across conditions (Pearson
$r \approx -0.3$ to $-0.5$ depending on task), but the relationship is
not strong enough to treat MC as a universal capacity metric.
Algorithm identity and connectome-specific structure together explain
the dominant fraction of residual variance: two conditions with identical
MC can differ substantially in NRMSE depending on which optimiser was used
and which species provided the reservoir topology.
This finding motivates the multi-task evaluation design of the paper: no
single task is sufficient to characterise reservoir quality.

\section*{S6.\enspace Optimiser Convergence Curves}

\begin{figure}[!ht]
\centerline{\includegraphics[width=\columnwidth]{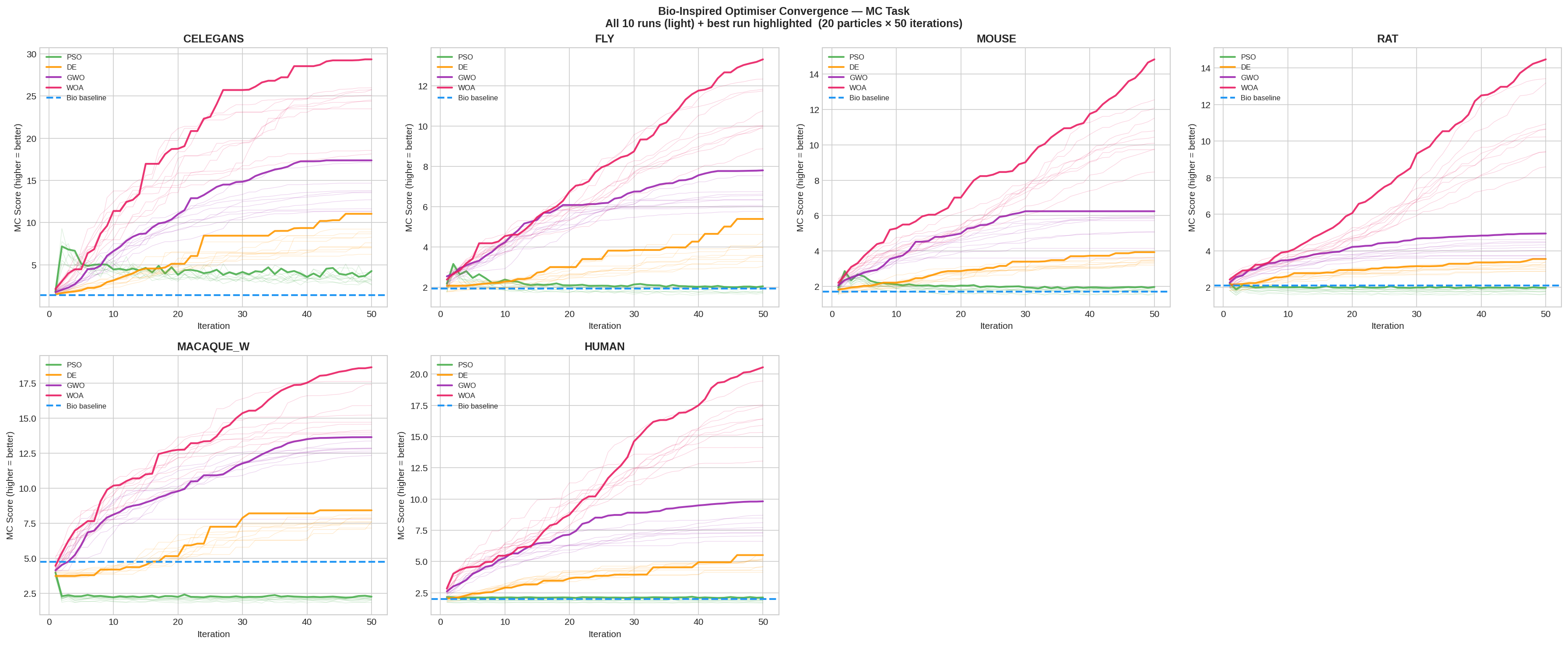}}
\caption{Convergence curves for the Memory Capacity (MC) task.
All 10 runs are shown; the bold line indicates the best run per species.
The dashed horizontal line marks the unoptimised biological baseline.}
\label{fig:convergence_mc}
\end{figure}

\begin{figure}[!ht]
\centerline{\includegraphics[width=\columnwidth]{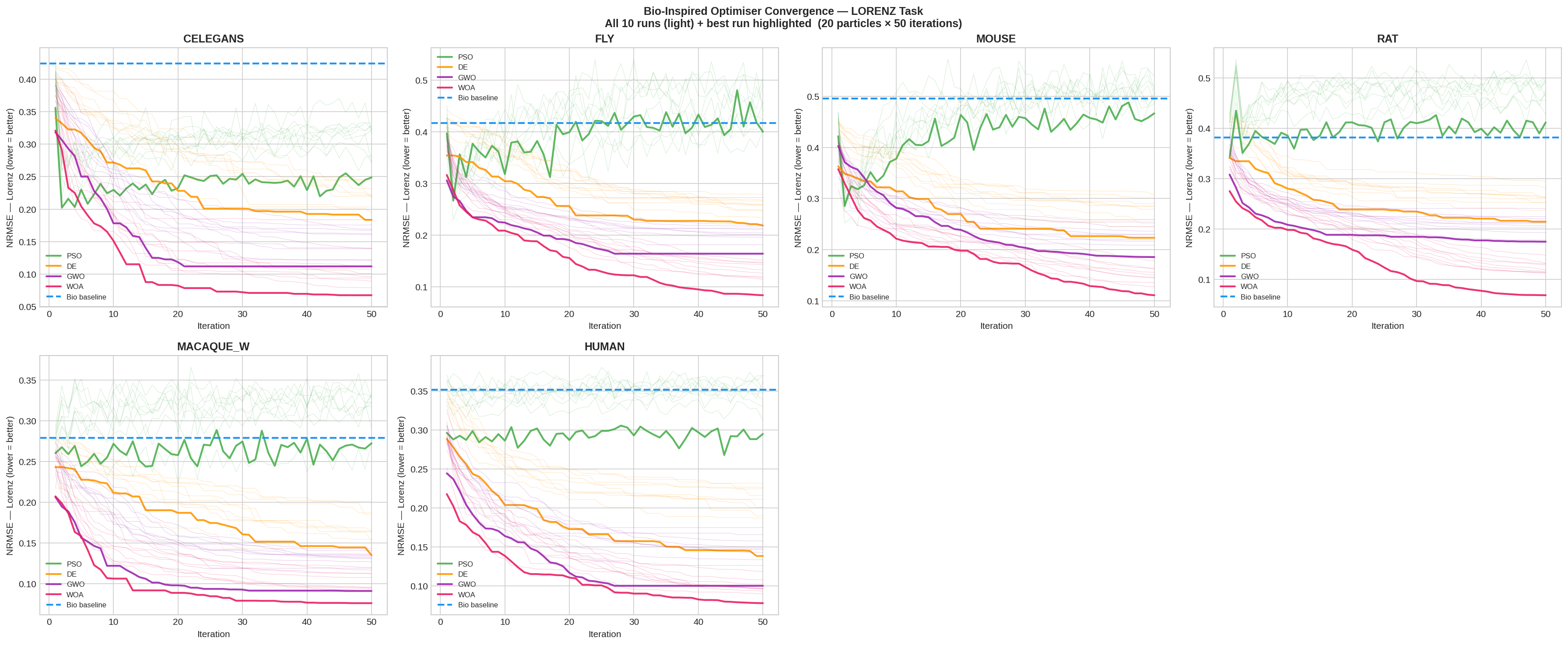}}
\caption{Convergence curves for the Lorenz attractor prediction task (NRMSE,
lower is better). All algorithms reduce NRMSE substantially relative to
the biological baseline; WOA achieves the lowest final values.}
\label{fig:conv_lorenz}
\end{figure}

\begin{figure}[!ht]
\centerline{\includegraphics[width=\columnwidth]{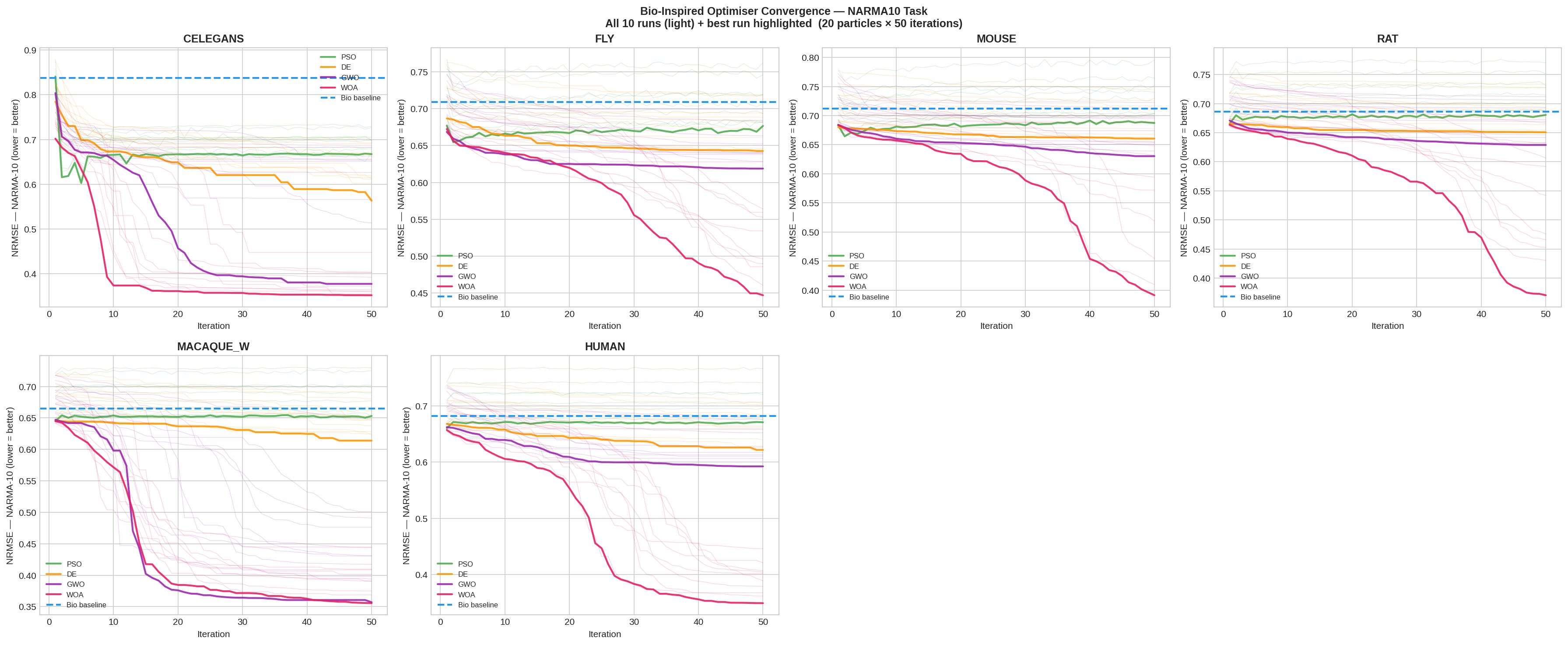}}
\caption{Convergence curves for the NARMA-10 task (NRMSE, lower is better).
Gains are more modest than on MC or Mackey--Glass, consistent with
NARMA-10's simultaneous demand for nonlinear computation and extended memory.}
\label{fig:conv_narma}
\end{figure}

\begin{figure}[!ht]
\centerline{\includegraphics[width=\columnwidth]{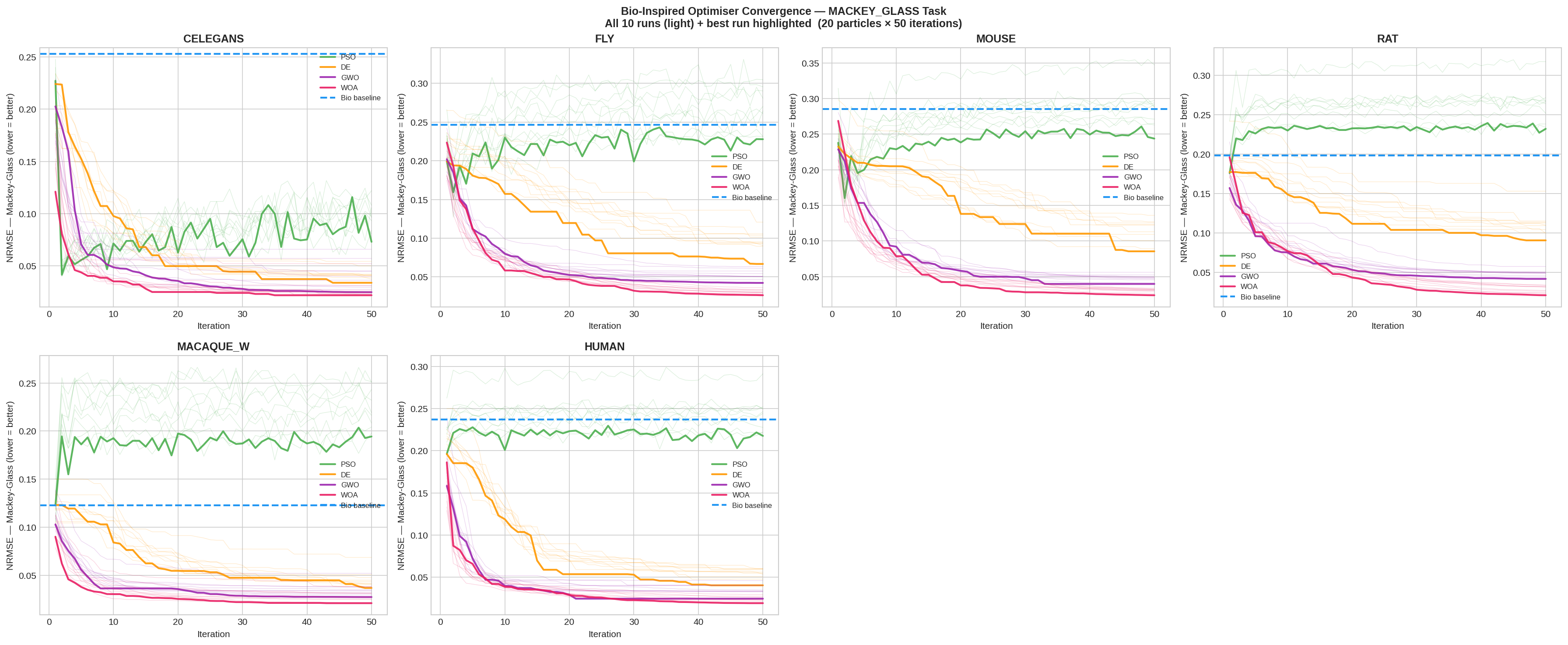}}
\caption{Convergence curves for the Mackey--Glass prediction task (NRMSE,
lower is better). WOA achieves near-elimination of NRMSE relative to
biology on several species, consistent with its dominance on this task
in the main results.}
\label{fig:conv_mg}
\end{figure}

Figs.~\ref{fig:convergence_mc}--\ref{fig:conv_mg} show fitness curves
across all 10 independent runs for each of the four benchmark tasks.
Bold lines mark the best run per species; the dashed horizontal line marks
the unoptimised biological baseline.

Three consistent patterns hold across all tasks and species.
First, all algorithms plateau well before iteration~50, confirming that
1{,}000 fitness evaluations constitute a sufficient budget for these
connectome sizes; no algorithm shows meaningful improvement in the final
10 iterations.
Second, WOA and GWO reach the highest final plateaus in every task, with
WOA's advantage most pronounced on MC and Mackey--Glass where the
exploration--exploitation balance of the bubble-net mechanism is
particularly effective.
Third, DE exhibits the lowest inter-run variance across all tasks, making
it the most reproducible choice when consistency across experimental
replications matters more than peak performance.
PSO converges most slowly and plateaus lowest, particularly on MC, reflecting
its sensitivity to the density of the weight landscape in high-dimensional
connectomes.

